\documentclass{article}
\usepackage{spconf,amsmath,graphicx}
\usepackage{fancyhdr}
 \usepackage{epsfig,amssymb,amsmath,multirow,boldline,graphicx,makecell,hyperref}
\usepackage{kotex}
\usepackage{color,soul}
\usepackage{multirow}
\usepackage{bigstrut}

\usepackage{xcolor}

\usepackage{color,soul}
\usepackage{multirow}
\usepackage{bigstrut}

\title{Efficient Arabic emotion recognition using deep neural networks}
\name{Yasser Hifny$^{\star}$, Ahmed Ali$^{\dagger}$ }
\address{$^{\star}$University of Helwan, Egypt\\
  $^{\dagger}$Qatar Computing Research Institute, HBKU, Doha, Qatar\\
  {\small \tt yhifny@fci.helwan.edu.eg \qquad amali@qf.org.qa } }

\begin{document}
%
\maketitle
\begin{abstract}
Emotion recognition from speech signal based on deep learning is an active research area. Convolutional neural networks (CNNs) may be the dominant method in this area. In this paper, we implement two neural architectures to address this problem.  The first architecture is an attention-based CNN-LSTM-DNN model. In this novel architecture, the convolutional layers extract salient features and the bi-directional long short-term memory (BLSTM) layers handle the sequential phenomena of the speech signal. This is followed by an attention layer, which extracts a summary vector that is fed to the fully connected dense layer (DNN), which finally connects to a softmax output layer. The second architecture is based on a deep CNN model. The results on an Arabic speech emotion recognition task show that our innovative approach can lead to significant improvements (2.2\% absolute improvements) over a strong deep CNN baseline system.  On the other hand, the deep CNN models are significantly faster than the attention based CNN-LSTM-DNN models in training and classification.
\end{abstract}

\section{Introduction}
Deep architectures for emotion recognition from speech is a growing research field \cite{anagnostopoulos2015features,schuller2018speech,neumann2017attentive,sarmaemotion,zhang2018attention}.  Using short time signal analysis, a speech utterance is represented by a matrix $X\in R^{d \times T}$ where $T$ is the size of the time dimension and  $d$ is the size of spectral dimension. The sequence to sequence layers 
model the spectral phenomena and keep the size of the time dimension $T$ without any modification.  A sequence to vector layer is used to convert the sequence to a fixed dimension vector which can be fed to feed forward dense layers. The global average pooling, global max pooling and attention are common choices for this type of layer. 
Fully-connected dense layers are then used to apply nonlinear compression of the input features for better representation which improves the modeling power of the classifier. A multiclass classifier is implemented using a softmax layer. Typically, the model is trained using the cross-entropy objective function.

Convolutional neural networks (CNNs) have been recently used in many emotion recognition tasks.  For example, CNNs designed for visual recognition (e.g. AlexNet) were directly adapted for emotion recognition from speech~\cite{zhang2018attention}. Moreover, in a study by \cite{neumann2017attentive}, they conducted extensive experiments using an attentive CNN with multi-view learning objective function using the Interactive Emotional Dyadic Motion Capture (IEMOCAP) database \cite{busso2008iemocap}. They concluded that for a CNN architecture, the particular choice of features is not as important as the model architecture, or the amount and kind of training data. 
CNNs are excellent in feature extraction and very fast in training compared to standard sequence modeling.

Long short-term memory networks (LSTMs) \cite{hochreiter1997long} sequence to sequence layers  are excellent in capturing the sequential phenomena of the speech signal for various style of speaking. In a study by \cite{trigeorgis2016adieu}, they propose a solution to the problem of ‘context-aware’ emotional relevant feature extraction, by combining CNNs with LSTM networks, in order to automatically learn the best representation of the speech signal directly from the raw time representation. They did not use any of the commonly hand-engineered features, such as mel-Frequency cepstral coefficients (MFCC) and perceptual linear prediction (PLP) coefficients. Their end-to-end system was targeted to learn an intermediate representation of the raw input signal automatically that better suits the task at hand and hence leads to better performance.

Both CNN and  LSTM networks have shown significant improvements over fully-connected neural network across a wide variety of tasks.  In recent work by  \cite{sainath2015convolutional},  they  took advantage of CNNs, LSTMs and DNNs by combining them into one unified architecture for speech recognition task. CNNs are good at reducing frequency variations, LSTMs are good at temporal modeling, and finally DNNs map the features into a more separable space. Their CLDNN provided a 4-6\% relative improvement in WER. In a similar work for emotion recognition from speech \cite{satt2017efficient}, the combination of CNNs and LSTMs led to improvements in the classification accuracy. The last state of the LSTM was used for sequence to vector conversion. 

Recently, an  end-to-end multimodal emotion and gender recognition model with dynamic joint loss weights is developed  \cite{chae2018end}. The proposed model does not need any pre-trained features from audio and visual data. In addition, the system is  trained using a multitask objective function and its weights are assigned using a dynamic approach.

In this paper, we build on these contributions to develop an emotion recognition system for Arabic data using the recently introduced KSU emotions corpus\footnote{https://catalog.ldc.upenn.edu/LDC97S45}. Our contributions are: (\textit{i}) Introducing a novel approach for emotion recognition by using an attention based CNN-LSTM-DNN architecture; (\textit{ii}) Studying a deep CNN models for the same task; (\textit{iii}) Comparing our results with published state-of-the art results on the IEMOCAP database and (\textit{iii}) Providing our scripts and code for  the research community for usage and potential future contributions\footnote{http://github.com/qcri/deepemotion}


The rest of the paper is organized as follows: In section 2, we describe the attention-based CNN-LSTM-DNN  and the proposed deep CNN architectures. Data is explained in section 3. Experimental setup is illustrated in section 4. This is followed by results in section 5. Finally section 6 concludes the paper and discusses future work.



\section{Proposed approach}

\begin{figure}[h]
\centering
 \includegraphics{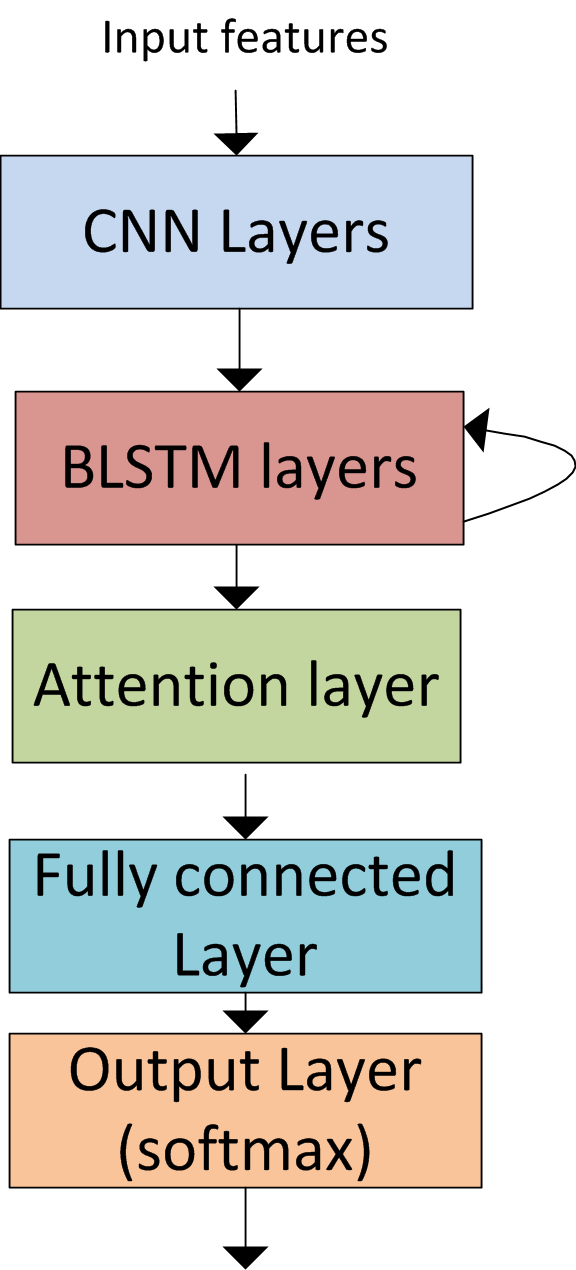} \\
  \caption{Attention based CNN-BLSTM-DNN architecture for speech emotion recognition.}\label{fig:CNN-LSTM-DNN}
\end{figure}

Our main emotion classifier is based on the CNN-LSTM-DNN  architecture as shown in Figure \ref{fig:CNN-LSTM-DNN}. The CNN layers  \cite{abdel2012applying}  extract features from the audio signal, while the bi-directional LSTM (BLSTM) layers  \cite{hochreiter1997long,graves2005bidirectional}  handle the sequential phenomena of the speech signal. This is followed by an attention layer, which extracts a summary vector that is fed to a DNN  layer, which finally connects to a softmax layer. 

\subsection{Attention layer}
The output of the BLSTM layer is followed by an attention layer, which extracts a summary vector that is fed to the fully connected dense layer (DNN), which finally connects to a softmax layer \cite{neumann2017attentive}. For each vector $\mathbf{x}_t$ in a sequence of inputs ${\mathbf{x}_1,\mathbf{x}_2,\ldots,\mathbf{x}_T}$, the attention weights $\alpha_t$  are given by:
\begin{equation}
\alpha_t=\frac{\exp (f(\mathbf{x}_t))}{\sum_{j=1}^T\exp (f(\mathbf{x}_j))}
\end{equation}
where $f(\mathbf{x}_t)$ is defined using the trainable parameters $\mathbf{w}$ as follows:
\begin{equation}
f(\mathbf{x}_t)=\tanh (\mathbf{w}^T\mathbf{x}_t)
\end{equation}

The summary vector which is output of the attention layer is computed as a weighted sum of the input sequence:
\begin{equation}
C=\sum_{t=1}^T \alpha_t\mathbf{x}_t
\end{equation}

\noindent Since most of the research in this area is based on CNN models \cite{neumann2017attentive,zhang2018attention}, we have implemented a strong CNN baseline.  

\section{Data} \label{sec:data}

We use the newly developed Arabic speech emotion database  (KSUEmotions) \cite{meftah2014designing} for our experiments. 
 The corpus was designed for Modern Standard Arabic (MSA) and was recorded using 23 speakers (10 males and 13 females).  The speakers are from three Arab countries: Yemen, Saudi Arabia and Syria. The speech was recorded in two phases. In Phase-1,  twenty speakers were selected (ten male speakers from Saudi Arabia, Yemen, and Syria and ten female speakers from two countries Saudi Arabia and Syria). In this phase, neutral, sadness, happiness, surprise, and questioning emotions were selected. 

Seven male speakers from phase one and seven female speakers (four of them from phase one and the other new three female speakers were added from Yemen nationality) were selected in phase two. In this phase, the questioning emotion was excluded while the anger emotion was added to this corpus to be more consistent with other similar corpora. The total duration of the all recorded files was 2 hours and 55 minutes for Phase-1 and 2 hours and 15 minutes for Phase-2. We mixed Phase-1 and  Phase-2 datasets to generate a bigger corpus (Phase-mix). Table \ref{tab:KSUEmotions} summarizes the statistics of this corpus. 

In an attempt to validate our findings, we built a strong baseline model using four emotional categories (neutral, angry, sad and happy) from the interactive emotional dyadic motion capture (IEMOCAP) database \cite {busso2008iemocap}. The database consists of about 12 hours of audiovisual data (speech, video, facial motion capture) from five mixed gender pairs of male and female actors, at two recording scenarios: scripted and improvised speech. The IEMOCAP corpus have five sessions, four of which are used for training and the remaining one is used for testing. Each wave file is manually labeled and has segment level emotion category.

\begin{table}
  \centering
  \caption{KSU emotions dataset statistics.}
    \begin{tabular}{|l|l|p{14.5em}|}
    \hline
    Phase & Duration & \multicolumn{1}{l|}{Supported emotions} \bigstrut\\
    \hline
    Phase1 & 3.0 hrs    & neutral, sadness, happiness, surprise, and questioning   \bigstrut\\
    \hline
    Phase2 & 2.2 hrs & neutral, sadness, happiness, surprise, and anger   \bigstrut\\
    \hline
        Mixed & 5.2 hrs & neutral, sadness, happiness, surprise, questioning, and anger   \bigstrut\\
    \hline
    \end{tabular}%
  \label{tab:KSUEmotions}%
\end{table}%

\section{Experimental setup}

\subsection{Input features}
We use the Essentia toolkit \cite{ bogdanov2013essentia} to  extract 13 MFCCs. The audio signal is segmented into 25 ms long frames with a 10 ms shift. A Hamming window is applied and the FFT with 512 points is computed. Then, we compute the logarithmic power of 26 Mel-frequency filter-banks over a range from 0 to 8 kHz. Finally, a discrete cosine transform (DCT) is applied to extract the first 13 MFCCs.  These features are used to train our emotion classifiers. 

\subsection {Training platform}

We implemented the emotion classification models using Keras deep learning library version 2.0 with Tensorflow 1.6.0~\cite{chollet2015keras} backend.  The  models were trained using an Nvidia’s Tesla K80 graphical processing unit (GPU). We used the  Adam optimization algorithm \cite{kingma2014adam} to train our models with $\beta_1=0.9$ and $\beta_2=0.999$. The initial learning rate was set to 0.001 and the  batch size was set to 32. Shorter sequences of a batch  were padded with zeros to match the longest sequence in the data (i.e. 15.5 seconds).

\subsection {Models}

A strong  deep CNN  emotion classifier  was developed as a baseline. It has six layers where four of them are CNN  layers (see Table \ref{tab:CNN_MODEL}). Dropout rate between layers was set to $0.2$.  We benchmarked our results with the well-known IEMOCAP database \cite{busso2008iemocap}.

Our experiment led to classification accuracy of 56.0\% which is similar to the state-of-the art results on this task~\cite{neumann2017attentive}. Hence, we used this model as a baseline to train  the Arabic emotion recognition task. Our CNN-LSTM-DNN emotion classifier has nine layers (see Table \ref{tab:Model}). Both models are trained using the crossentropy objective function.

\begin{table}[htbp]
  \centering
  \caption{The deep CNN architecture.}
    \begin{tabular}{|r|l|p{14.5em}|}
    \hline
    \multicolumn{1}{|l|}{Layer} & Type  & \multicolumn{1}{l|}{Details} \bigstrut\\
    \hline
    1     & Conv  & 500 filters + Relu + Stride=1 + kernal wdith=5 \bigstrut\\
    \hline
    2     & Conv  & 500 filters + Relu + Stride=2 + kernal wdith=7 \bigstrut\\
    \hline
    3     & Conv  & 500 filters + Relu + Stride=2 + kernal wdith=1 \bigstrut\\
    \hline
    4     & Conv  & 500 filters + Relu + Stride=1 + kernal wdith=1 \bigstrut\\
    \hline
    5     & MaxPool1D & Global \bigstrut\\
    \hline
    6     & Softmax & 6 classes for Phase-mix dataset \bigstrut\\
    \hline
    \end{tabular}%
  \label{tab:CNN_MODEL}%
\end{table}%

\begin{table}[bt]
  \centering
  \caption{The CNN-LSTM-DNN architecture.}
    \begin{tabular}{|r|l|p{14.5em}|}
    \hline
    \multicolumn{1}{|l|}{Layer} & Type  & Details \bigstrut\\
    \hline
    1     & Conv  & 256 filters + Relu + Stride=1 + kernal wdith=5  \bigstrut\\
    \hline
    2     & MaxPool1D & Pool size=2 \bigstrut\\
    \hline
    3     & Conv  & 64 filters + Relu + Stride=1 + kernal wdith=5 \bigstrut\\
    \hline
    4     & MaxPool1D & Pool size = 2 \bigstrut\\
    \hline
    5     & BLSTM &  128 units (64 in each direction) \bigstrut\\
    \hline
    6     & BLSTM &  128 units (64 in each direction) \bigstrut\\
    \hline
    7     & Attention & 128 dimension \bigstrut\\
    \hline
    8     & Dense & 64 units + tanh activation \bigstrut\\
    \hline
    9     & Softmax & 6 classes for Phase-mix dataset \bigstrut\\
    \hline
    \end{tabular}%
  \label{tab:Model}%
\end{table}%

\section{Results}

\begin{figure*}[t]
    \centering
    \begin{minipage}{0.33\textwidth}
        \centering
        \includegraphics[width=1\textwidth]{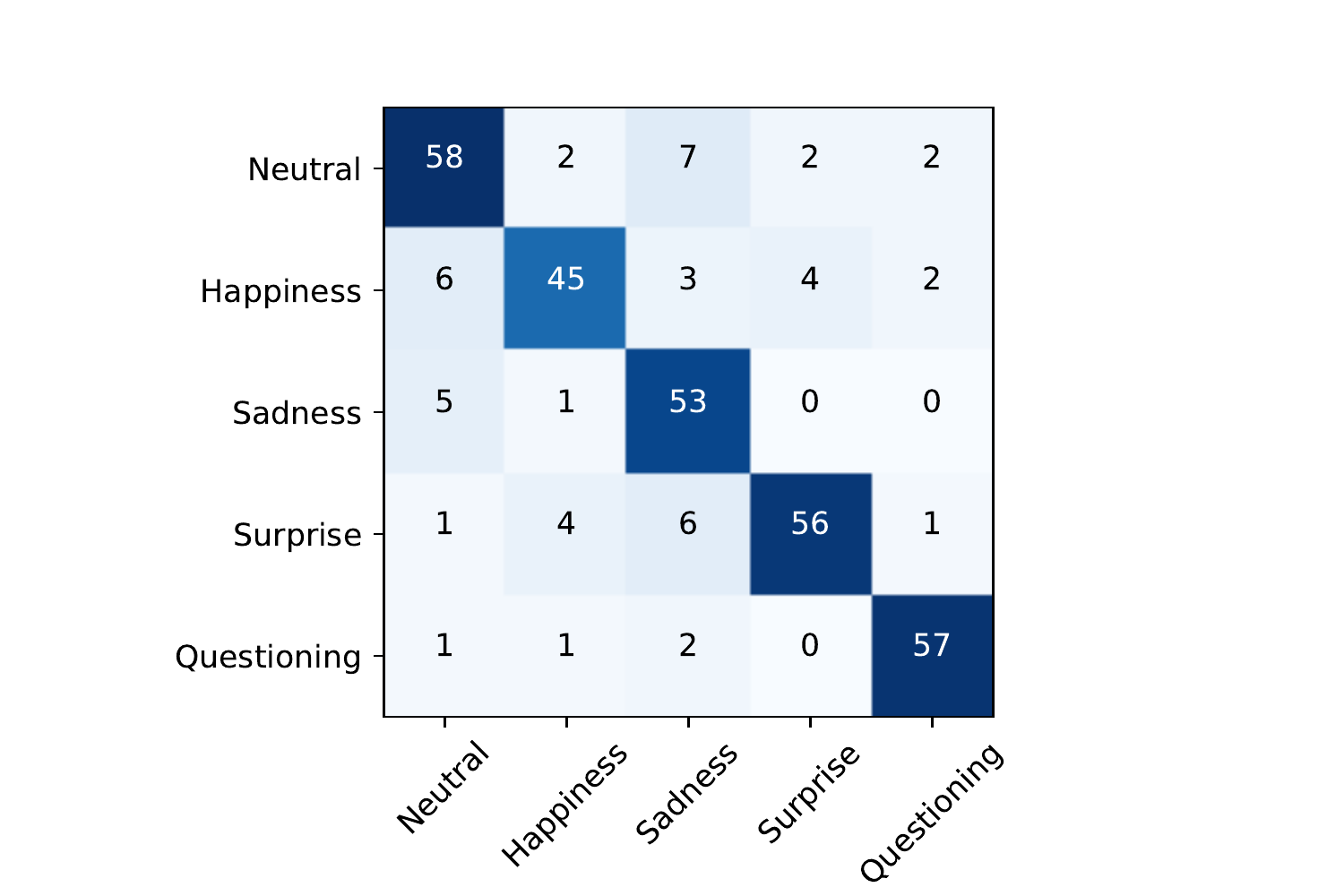}
        \label{ch:phase1_cm}
        (a) Confusion matrix for Phase-1.
    \end{minipage}\hfill
    \begin{minipage}{0.33\textwidth}
        \centering
        \includegraphics[width=1\textwidth]{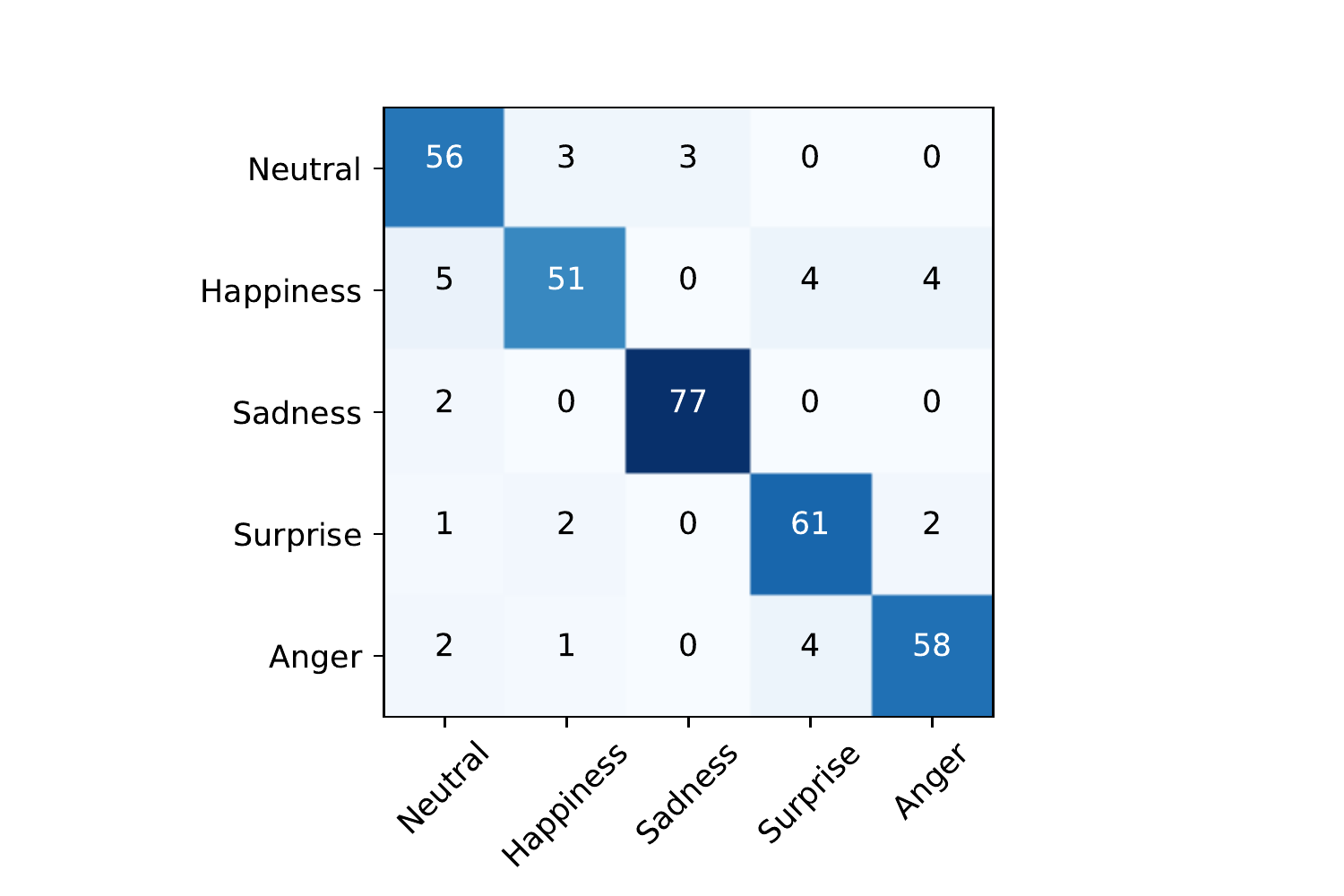}
        \label{ch:phase2_cm}
        (b) Confusion matrix for Phase-2.
    \end{minipage}
    \begin{minipage}{0.33\textwidth}
        \centering
        \includegraphics[width=1\textwidth]{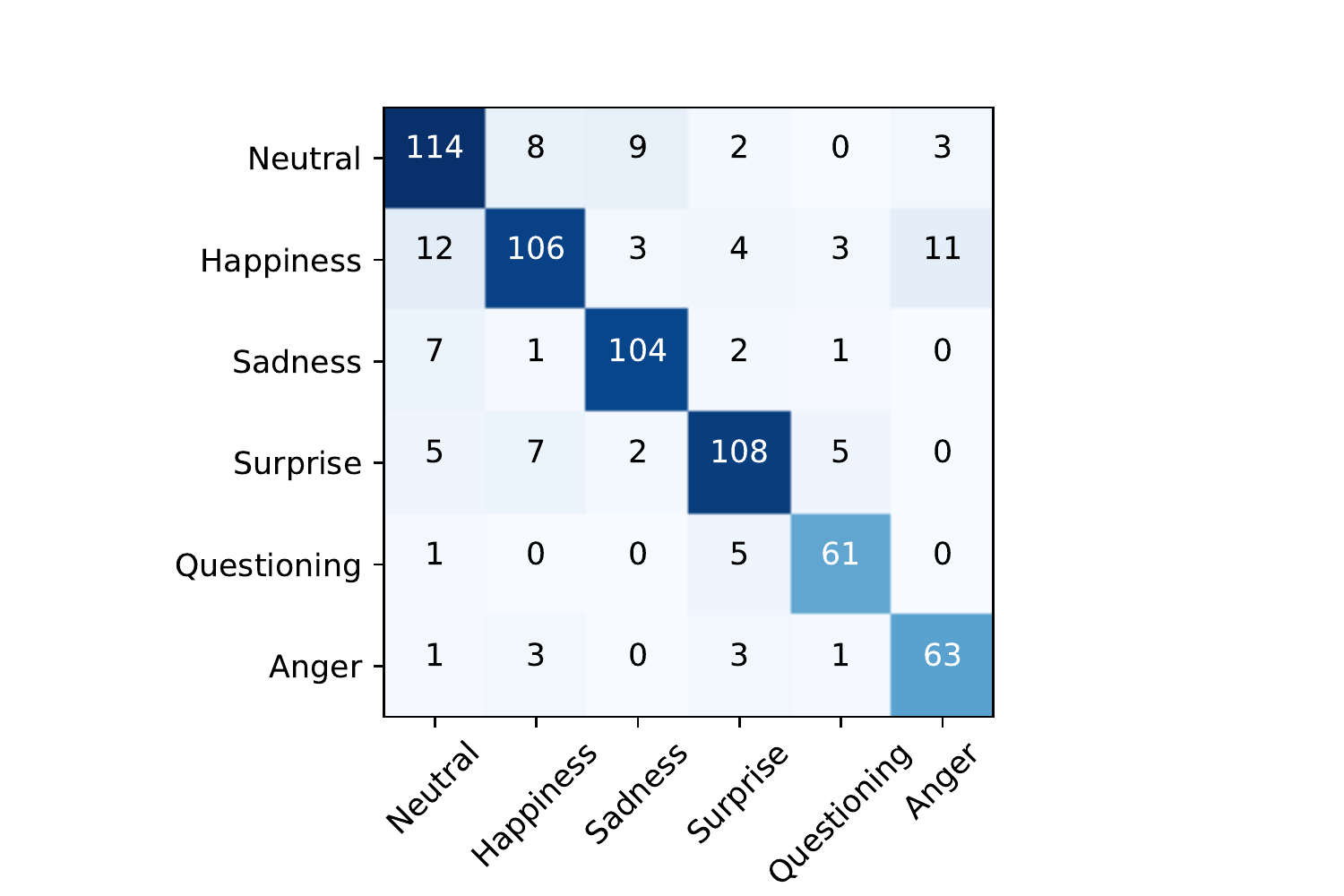}
        \label{ch:phase2_cm}
        (c) Confusion matrix for Phase-mix.
    \end{minipage}
     \caption{Confusion matrix for different phases. Note that the $x$ axis shows predictions and the $y$ axis shows the labels. }
     \label{fig:cm}
\end{figure*}

\begin{figure*}[t]
    \centering
    \begin{minipage}{0.33\textwidth}
        \centering
        \includegraphics[width=1.1\textwidth]{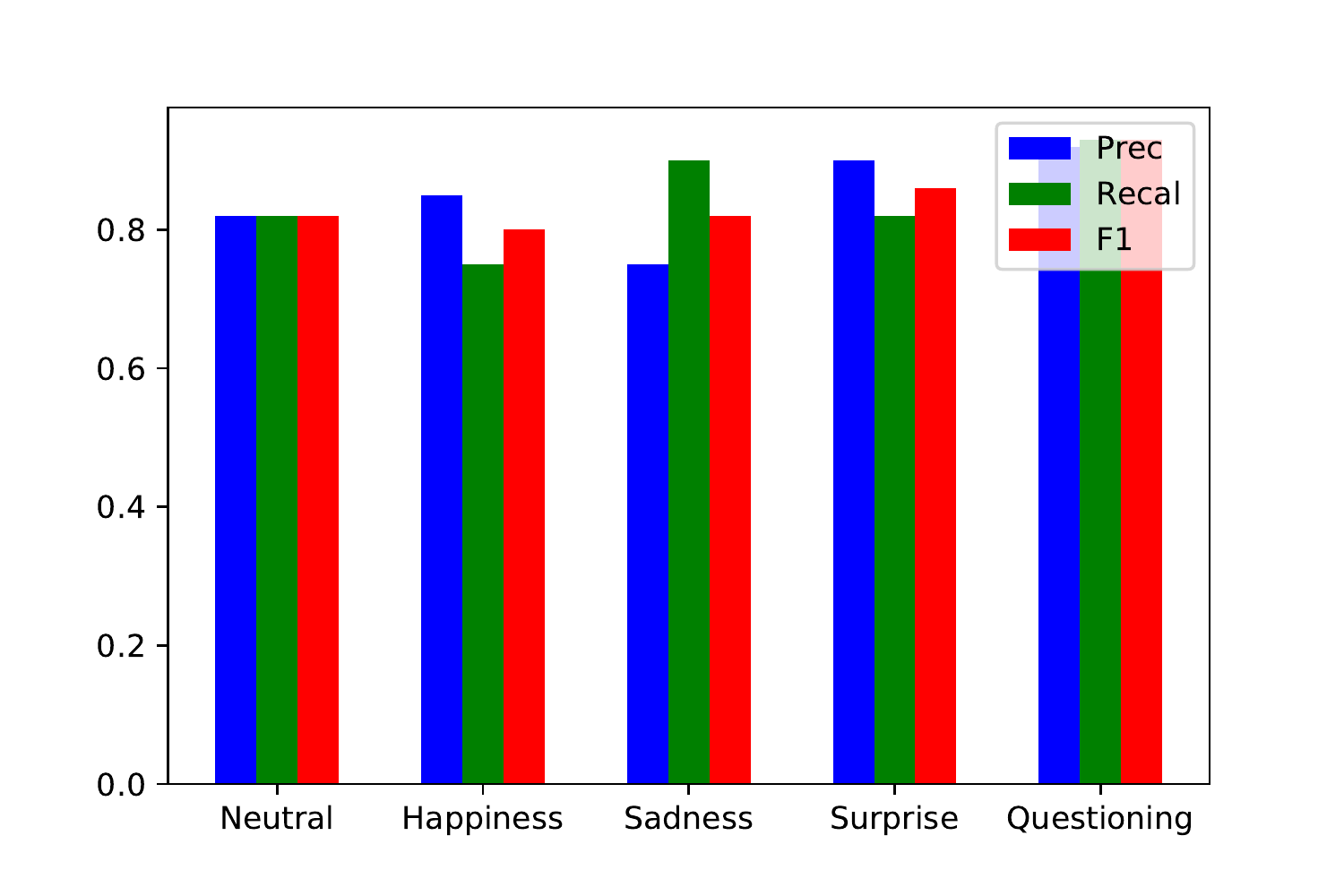}
        \label{ch:phase1_cm}
        (a) The F1 results for Phase-1 dataset.
    \end{minipage}\hfill
    \begin{minipage}{0.33\textwidth}
        \centering
        \includegraphics[width=1.1\textwidth]{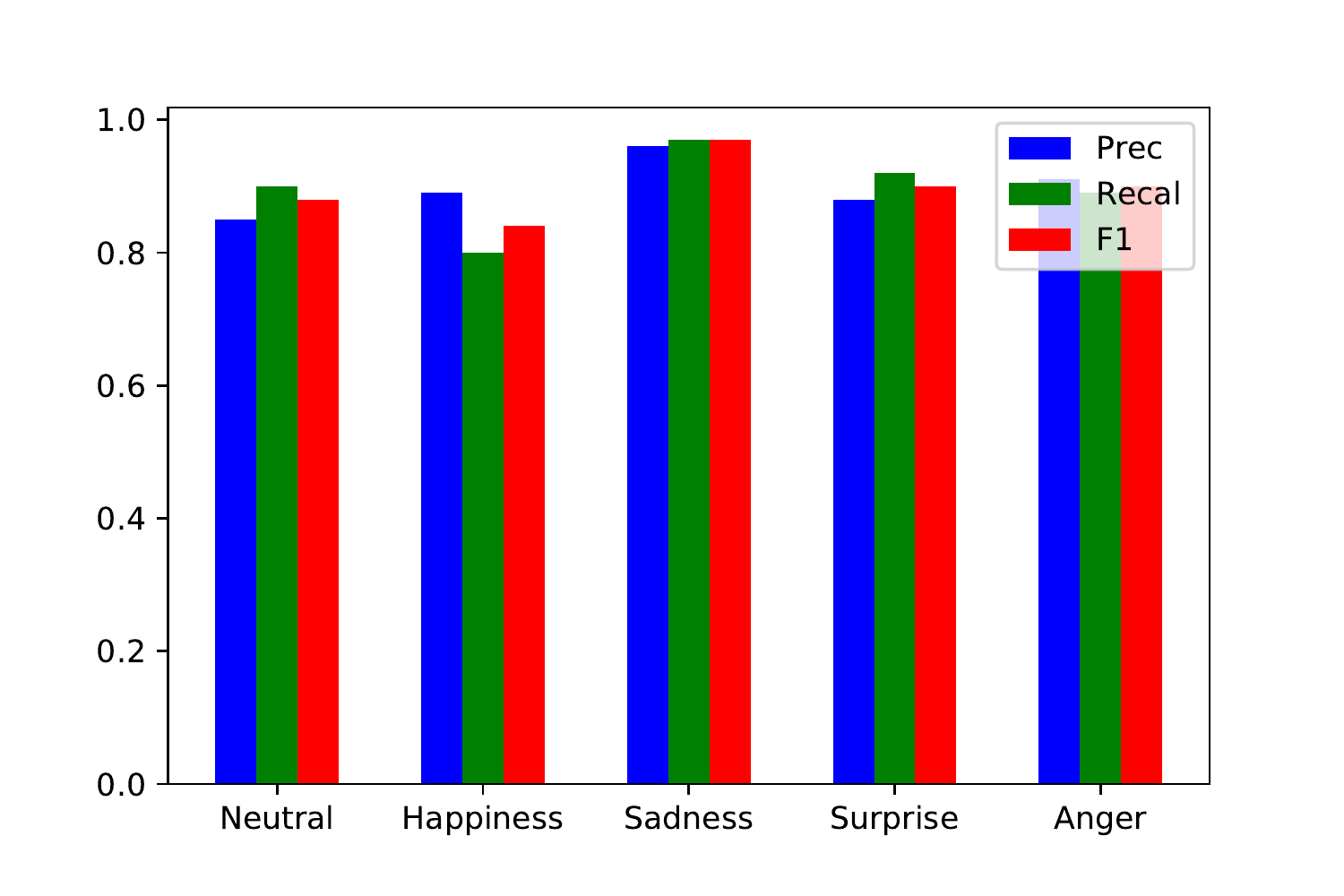}
        \label{ch:phase2_cm}
        (b) The F1 results for Phase-2 dataset.
    \end{minipage}
    \begin{minipage}{0.33\textwidth}
        \centering
        \includegraphics[width=1.1\textwidth]{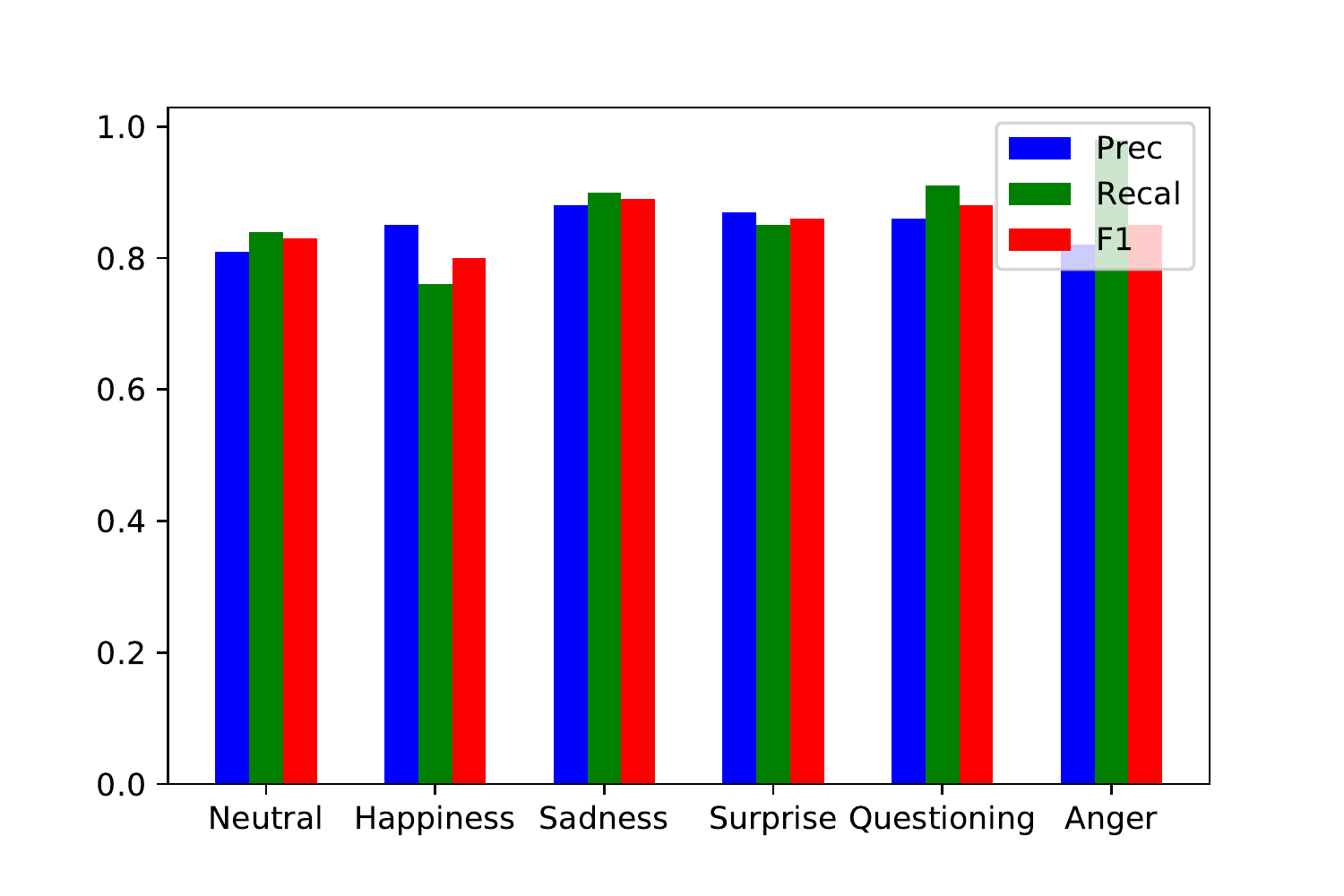}
        \label{ch:phase2_mix}
        (c) The F1 results for Phase-mix dataset.
    \end{minipage}
    \caption{Overall F1 results for all the different emotion classes.}
    \label{fig:f1}
\end{figure*}

Our results were generated using Phase-1, Phase-2 and Phase-mix datasets. 
 We ran five folds cross validation experiments
, the results over the five folds were consistent. Hence, we report the results for fold number five experiments for each phase. In Figure \ref{fig:cm}, the confusion matrices for each experiment are detailed. We observe that in Phase-1 experiment,  the questioning and sadness emotions were recognized successfully most of the time.  In Phase-2, sadness emotion was confused less than other emotions. In Phase-mix, the classification problem seems to be harder than Phase-1 and Phase-2 datasets. The overall precision, recall and F1 scores for each emotion are shown in Figure \ref{fig:f1}. It is worth noting that both emotions, questioning and anger performed  poorly compared to the rest of the emotions in the Phase-mix as shown in Figure \ref{ch:phase2_mix}. This is primarily due to less data being used for these two emotions in the combined dataset.



Table \ref{tab:COMP} shows the results for both the  attention-based CNN-BLSTM-DNN and deep CNN systems. The attention-based CNN-BLSTM-DNN  model  results are consistently  better than the  deep CNN model. When the attention layer is replaced with last state of the  second LSTM layer as in \cite{satt2017efficient}, the classification accuracy of the model drops  significantly (i.e. more than 2\% absolute). On the other hand, the CNN baseline is much faster than the attention-based CNN-BLSTM-DNN models in training and classification.

\begin{table}[htbp]
  \centering
  \caption{Classification results using the Phase-mix dataset.}
    \begin{tabular}{|l|l|l|}
    \hline
    Fold  & \multicolumn{2}{c|}{Classification overall accuracy in \%} \bigstrut\\
    \hline
          & CNN-BLSTM-DNN & CNN \bigstrut\\
    \hline  \hline
    1     & 87.7  & 85.2 \bigstrut\\
    \hline
    2     & 86.6  & 85.6 \bigstrut\\
    \hline
    3     & 88.2    & 84.4 \bigstrut\\
    \hline
    4     & 86.0       & 85.2 \bigstrut\\
    \hline
    5     & 87.6      & 84.7 \bigstrut\\
    \hline 
        \hline 
    Average     & 87.2      & 85.0 \bigstrut\\
    \hline        
    \end{tabular}%
  \label{tab:COMP}%
\end{table}%

\section{Conclusions}

In this paper, we designed an end-to-end attention-based CNN-LSTM-DNN emotion classifier.   In our classifier, the convolutional layers (CNN) extract salient features, the bi-directional long short-term memory (BLSTM) layers handle the sequential phenomena of the speech signal. This is followed by an attention layer, which extracts a summary vector that is fed to the fully connected dense layer (DNN), which finally connects to a softmax layer. The results on an Arabic speech emotion recognition task show that our innovative approach can lead to significant improvements (2.2\% absolute improvements) over a strong deep CNN baseline system.  However, the deep CNN models are significantly faster than the attention-based CNN-LSTM-DNN models in training and classification phases. 

Future work will focus on training an ensemble classifier and interpolating the predictions to improve the classification accuracy. We plan to use large Arabic emotion databases using our powerful attention-based CNN-LSTM-DNN models. In addition, joint estimation of the emotion, dialect, language, and accent using multitask learning will be investigated.   In addition, the  separate label per  frame  methods developed in~\cite{sarmaemotion} will be compared with  the  single label per utterance methods commonly used in the field in a unified framework.

\pagebreak

\bibliographystyle{IEEEbib}
\bibliography{refs}

\end{document}